\documentclass[runningheads]{llncs}

\usepackage{epsfig}
\usepackage{graphicx}
\usepackage{tabularx}
\usepackage{amsmath,mathtools}
\usepackage{amssymb}
\usepackage[inline]{enumitem}
\usepackage[dvipsnames,table]{xcolor}
\usepackage{booktabs}
\usepackage{tabularx}
\usepackage{multirow}
\usepackage{colortbl}
\usepackage{subcaption}
\usepackage{bbm}
\usepackage{caption}
\usepackage{fp}
\usepackage{diagbox}

\usepackage{algpseudocode,algorithm,algorithmicx}
\usepackage{xspace}
\usepackage{comment}
\usepackage{bm,upgreek}
\usepackage[pagebackref=true,breaklinks=true,letterpaper=true,colorlinks,citecolor=ForestGreen, bookmarks=false]{hyperref}

\newcommand{\enstwo}{{\scshape Ens2}\xspace}
\newcommand{\ensthree}{{\scshape Ens3}\xspace}
\newcommand{\ft}{{\scshape Fted}\xspace}
\newcommand{\yt}{{\scshape Yolov2}\xspace}

\makeatletter\renewcommand{\paragraph}{%
  \@startsection{paragraph}{4}{\z@}%
                {0.1em \@plus 0.5ex \@minus 0.3ex}%
                {-0.3em}%
                {\normalfont\normalsize\bf}%
}\makeatother

\DeclareMathOperator*{\minimize}{minimize}
\newcommand{\ra}[1]{\renewcommand{\arraystretch}{#1}}

\newcommand*{\eg}{\emph{e.g.}\@\xspace}
\newcommand*{\ie}{\emph{i.e.}\@\xspace}
\newcommand*{\etal}{\emph{et al.}\@\xspace}

\makeatletter
\newcommand*{\etc}{%
    \@ifnextchar{.}%
        {\emph{etc}}%
        {\emph{etc.}\@\xspace}%
}

\newcommand{\cval}[1]{ \cellcolor{SkyBlue!#1} #1  }

\newcommand{\render}{\mathcal R_\theta}

\graphicspath{{./figures/}}

\begin{document}
\pagestyle{headings}
\mainmatter
\def\ECCVSubNumber{1425}  

\title{Making an Invisibility Cloak: Real World Adversarial Attacks on Object Detectors} 


\titlerunning{Making an Invisibility Cloak}
%
\author{Zuxuan Wu\inst{1,2} \and
Ser-Nam Lim\inst{2} \and
Larry S. Davis\inst{1} \and {Tom Goldstein}\inst{1,2}}
\authorrunning{Wu Z. et al.}
%
\institute{$^1$University of Maryland, College Park~~~$^2$Facebook AI}

\maketitle

\begin{abstract}
 We present a systematic study of the transferability of adversarial attacks on state-of-the-art object detection frameworks.
Using standard detection datasets, we train patterns that suppress the objectness scores produced by a range of commonly used detectors, and ensembles of detectors. Through extensive experiments, we benchmark the effectiveness of adversarially trained patches under both white-box and black-box settings, and quantify transferability of attacks between datasets, object classes, and detector models. Finally, we present a detailed study of physical world attacks using printed posters and wearable clothes, and rigorously quantify the performance of such attacks with different metrics.
\end{abstract}

\begin{center}
\centering
\includegraphics[width=1.0\linewidth]{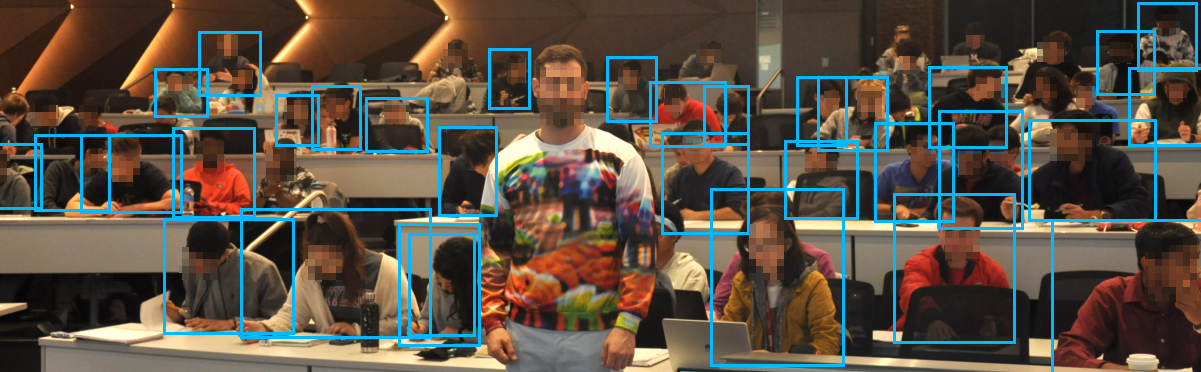}
\captionof{figure}{
In this demonstration, the YOLOv2 detector is evaded using a pattern trained on the COCO dataset with a carefully constructed objective.}
\label{fig:teaser}
\end{center}

\section{Introduction}
Adversarial examples are security vulnerabilities of machine learning systems in which an attacker makes small or unnoticeable perturbations to system inputs with the goal of manipulating system outputs. 
These attacks are most effective in the digital world, where attackers can directly manipulate image pixels.  However, many studies assume a white box threat model, in which the attacker knows the dataset, architecture, and model parameters used by the victim. In addition, most attacks have real security implications only when they cross into the physical realm. 

In a ``physical'' attack, the adversary modifies a real-world {\em object}, rather than a digital image, so that it confuses systems that observe it. These objects must maintain their adversarial effects when observed with different cameras, resolutions, lighting conditions, distances, and angles.

While a range of physical attacks have been proposed in the literature, these attacks are frequently confined to digital simulations,
or are demonstrated against simple classifiers rather than object detectors.
However, in most realistic situations the attacker has only black or grey box knowledge, and their attack must {\em transfer} from the digital world into the physical world, from the attacker model to the victim model, or from models trained on one dataset to another.  

In this paper, we study the transferability of attacks on object detectors across different architectures, classes, and datasets, with the ultimate goal of generalizing digital attacks to the real-world.  Our study has the following goals:

\begin{itemize}
\item We focus on industrial-strength detectors under both black-box and white-box settings.  
 Unlike  {\em classifiers}, which output one feature vector per image, object {\em detectors} output a map of vectors, one for each prior (\ie, candidate bounding box), centered at each output pixel.  Since any of these priors can detect an object, attacks must simultaneously manipulate hundreds or thousands of priors operating at different positions, scales, and aspect ratios.
\item In the digital setting, we systematically quantify how well attacks on detectors transfer between models, classes and datasets.
\item We break down the incremental process of getting attacks out of a digital simulation and into the real world.  We explore how real-world nuisance variables cause major differences between the digital and physical performance of attacks, and present experiments for quantifying and identifying the sources of these differences.
\item We {\em quantify} the success rate of attacks under various conditions, and measure how algorithm and model choices impact success rates.  We rigorously study how attacks degrade classifiers using standard metrics (average precision) that best describe the strength of detectors, and also more interpretable success/failure metrics.   
\item We push physical attacks to their limits with wearable adversarial clothing, and quantify the success rate of our attacks under complex fabric distortions. 
\end{itemize}

\section{Related Work}
\paragraph{Attacks on object detection and semantic segmentation.} While there is a plethora of work on attacking image classifiers~\cite{moosavi2017universal,goodfellow2014explaining,madry2017towards}, less work has been done on more complex vision tasks like object detection and semantic segmentation. Metzen \etal demonstrate that nearly imperceptible adversarial perturbations can fool segmentation models to produce incorrect outputs~\cite{metzen2017universal}. Arnab \etal also show that segmentation models are vulnerable to attacks~\cite{arnab2018robustness}, and claim that adversarial perturbations fail to transfer across network architectures. Xie \etal introduce Dense Adversary Generation (DAG), a method that produces incorrect predictions for pixels in segmentation models or proposals in object detection frameworks~\cite{xie2017adversarial}. Wei \etal further extend the attack from images to videos~\cite{wei2019transferable}. In contrast to~\cite{xie2017adversarial,wei2019transferable}, which Attacks the classifier stage of object detectors, Li \etal attack region proposal networks by decreasing the confidence scores of positive proposals~\cite{DBLP:conf/bmvc/LiTCBL18}. DPatch causes misclassification of detectors, by placing a patch that does not overlap with the objects of interest~\cite{liu2018dpatch}. Li \etal add imperceptible patches to the background to fool object detectors~\cite{li2018exploring}. Note that all of these studies focus on digital (as opposed to physical) attacks with a specific detector, without studying the transferability of attacks. In this paper, we systematically evaluate a wide range of popular detectors in both the digital and physical world, and benchmark how attacks transfer in different settings.

\paragraph{Physical attacks in the real world.} Kurakin \etal took photos of adversarial images with a camera and input them to a pretrained classifier~\cite{kurakin2016adversarial}; they demonstrate that a large fraction of images are misclassified. 
Eykholt \etal consider physical attacks on stop sign classifiers using images cropped from video frames~\cite{eykholt2017robust}.
They successfully fool classifiers using both norm bounded perturbations, and also sparse perturbations using carefully placed stickers. Stop signs attacks on object detectors are considered in~\cite{eykholt2018physical,chen2018shapeshifter}. 
Lu \etal showed that the perturbed sign images from~\cite{eykholt2017robust} can be reliably recognized by popular detectors like Faster-RCNN~\cite{ren2015faster} and Yolov2~\cite{redmon2017yolo9000}, and showed that detectors are much more robust to attacks than classifiers.  Note that fooling stop sign detectors differs from fooling person detectors because stop sign perturbations can cover the
whole object, whereas our person attacks leave the face, hands, and legs uncovered.
 
Zeng \etal use rendering tools to perform attacks in 3D environments~\cite{zeng2019adversarial}. Sitawarin \etal \cite{sitawarin2018darts} propose large out-of-distribution perturbations, producing toxic signs to deceive autonomous vehicles.  Athalye \etal introduce expectation over transformations (EoT) to generate physically robust adversarial samples, and they produce 3D physical adversarial objects that can attack classifiers in different conditions. Sharif \etal explore adversarial eyeglass frames that fool face classifiers~\cite{sharif2016accessorize}. Brown \etal placed adversarial patches \cite{brown2017adversarial} on raw images, forcing classifiers to output incorrect predictions.  Komkov \etal generate stickers attached to hats to attack face classifiers~\cite{komkov2019advhat}.  Huang \etal craft attacks by simulations to cause misclassification of detectors~\cite{huang2020universal}.
Thys \etal produce printed adversarial patches~\cite{thys2019fooling} that deceive person detectors instantiated by Yolov2~\cite{redmon2017yolo9000}.  This proof-of-concept study was the first to consider physical attacks on detectors, although it was restricted to the white-box setting.  Furthermore the authors did not quantify the performance, or address issues like robustness to distance/distortions and detectors beyond Yolov2, and the transferability of their attacks are limited. Xu \etal learn TPS transformations to generate T-shirts~\cite{xu2019adversarial}. To the best of our knowledge, no paper has conclusively demonstrated the transferability of attacks on object detectors, or quantified the reliability of transfer attacks.

\subsection{Object detector basics} 
We briefly review the inner workings of object detectors~\cite{lin2017focal,cascade,sniper,trident}, most of which can be described as two-stage frameworks (\eg, Fast(er) RCNN~\cite{girshick2015fast,ren2015faster}, Mask RCNN~\cite{he2017mask}, \etc) or one-stage pipelines (\eg, YOLOv2~\cite{redmon2017yolo9000}, SSD~\cite{liu2016ssd}, \etc).

\paragraph{Two-stage detectors.} 
These detectors use a region proposal network (RPN) to identify potential bounding boxes (Stage I), and then classify the contents of these bounding boxes (Stage II). 
An RPN passes an image through a {\em backbone} network to produce a stack of 2D feature maps with resolution $W' \times H'$ (or a feature pyramid containing features at different resolutions). The RPN considers $k$ ``priors'', or candidate bounding boxes with a range of aspect ratios and sizes, centered on every output pixel.  For each of the  $W' \times H'\times k$ priors, the RPN produces an ``objectness score'', and also the offset to the center coordinates and dimensions of the prior to best fit the closest object.
Finally, proposals with high objectness scores are sent to a Stage-II network for classification.

\paragraph{One-stage detectors.} These networks generate object proposals and at the same time predict their class labels. Similar as RPNs, these networks typically transform an image into a $W' \times H'$ feature map, and each pixel on the output contains the locations of a set of default bounding boxes, their class prediction scores, as well as objectness scores.

\paragraph{Why are detectors hard to fool?} A detector usually produces hundreds or thousands of priors that overlap with an object.   Usually, non-maximum supression (NMS) is used to select the bounding box with highest confidence, and reject overlapping boxes of lower confidence so that an object is only detected once.  Suppose an adversarial attack evades detection by one prior.  In this case, the NMS will simply select a different prior to represent the object.  For an object to be completely erased from an image, the attack must simultaneously fool the ensemble of all priors that overlap with the object---a much harder task than fooling the output of a single classifier.

\paragraph{Detection metrics.} In the field of object detection, the standard performance metric is average precision (AP) per class, which balances the trade-off between precision and recall. In contrast, success rates (using a fixed detection threshold) are more often used when evaluating physical attacks, due to their interpretability. However, manually selected thresholds are required to compute success rates. To mitigate this issue, we report both AP (averaging over all confidence thresholds) and success rates.

\section{Approach}
Our goal is to generate an adversarial pattern that, when placed over an object either digitally or physically, makes that object invisible to detectors. Further, we expect the pattern to be (1) universal (image-agnostic)---the pattern must be effective against a range of objects and within different scenes; (2) transferable---it breaks a variety of detectors with different backbone networks; (3)  dataset agnostic---it should fool detectors trained on disparate datasets;  (4) robust to viewing conditions---it can withstand field-of-view changes when observed from different perspectives and distances; (5) realizable---patterns should remain adversarial when printed over real-world 3D objects.  

\begin{figure}[h!]
    \centering 
    \includegraphics[width=0.9\linewidth]{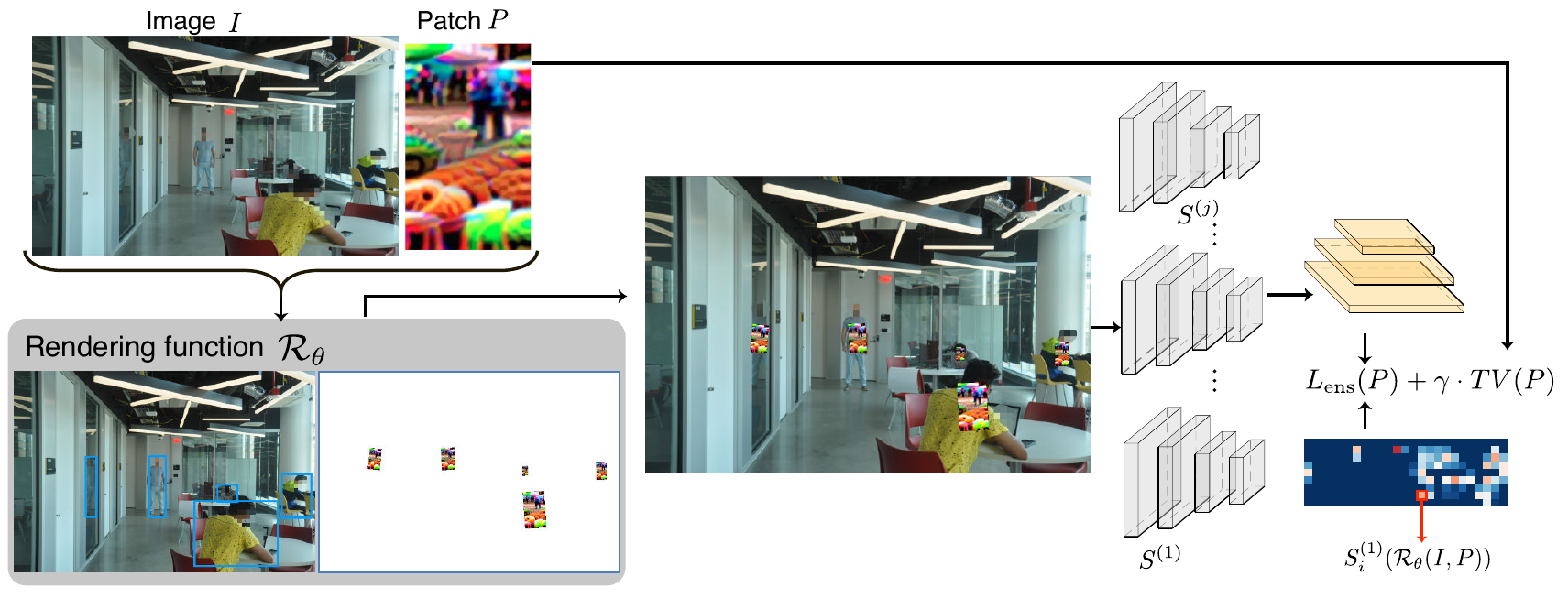}
    \caption{\textbf{An overview of the framework}. Given a patch and an image, the rendering function uses translations and scalings, plus random augmentations, to overlay the patch onto detected persons. The patch is then updated to minimize the objectness scores produced by a detector while maintaining its smoothness. }
    \label{fig:framework}
\end{figure}

\subsection{Creating a universal adversarial patch}
Our strategy is to ``train'' a patch using a large set of images containing people.  On each iteration, we draw a random batch of images, and pass them through an object detector to obtain bounding boxes containing people.  We then place a randomly transformed patch over each detected person, and update the patch pixels to minimize the objectness scores in the output feature map. 

More formally, we consider a patch $P \in \mathbb{R}^{w \times h \times 3}$ and a randomized rendering function $\render$.  The rendering function takes a patch $P$ and image $I,$ and renders a rescaled copy of $P$ over every detected person in the image $I.$  In addition to scaling and translating the patch to place it into each bounding box, the rendering function also applies an augmentation transform parameterized by the (random) vector $\theta.$

These transforms are a composition of brightness, contrast, rotation, translation, and sheering transforms that help make patches robust to variations caused by lighting and viewing angle that occur in the real world.  We also consider more complex thin-plate-spline (TPS) transforms to simulate the random ``crumpling'' of fabrics. 

A detector network takes a patched image $\render(I,P)$ as its input, and outputs a vector of objectness scores, $\mathcal S(\render(I,P))$ one for each prior.  These scores rank general objectness for a two-stage detector, and the strength of the ``person'' class for a one-stage detectors. A positive score is taken to mean that an object/person overlaps with the corresponding prior, while a negative score denotes the absence of a person.  To minimize objectness scores, we use the objectness loss function
\begin{align}
\label{eqn:loss_obj}
L_\text{obj}(P) = \mathbb{E}_{\theta,I} \sum_{i}\max \{\mathcal  S_i(\render(I,P))+1, \, 0\}^2.
\end{align}
Here, $i$ indexes the priors produced by the detector's score mapping.  The loss function penalizes any objectness score greater than $-1$.  This suppresses scores that are positive, or lie very close to zero, without wasting the ``capacity'' of the patch on decreasing scores that are already far below the standard detection threshold.  We minimize the expectation over the transform vector $\theta$ as in~\cite{athalye2017synthesizing}  to promote robustness to real-world distortions, and also the expectation over the random image $I$ drawn from the training set.

Finally, we add a small total-variation penalty to the patch.  We do this because there are pixels in the patch that are almost never used by the rendering function $\render,$ which down-samples the high-resolution patch (using linear interpolation) when rendering it onto the image.  A small TV penalty ensures a smooth patch in which all pixels in the patch get optimized.  A comparison to results without this TV penalty is shown in the supplementary material.  The final optimization problem we solve is
\begin{align}
    \label{eqn:loss}
    \minimize_{P}   L_\text{obj}(P) + \gamma \cdot  TV(P),
\end{align}
where $\gamma$ was chosen to be small enough to prevent outlier pixels without visibly distorting the patch.

\paragraph{Ensemble training.} To help adversarial patterns generalize to detectors that were not used for training (\ie, to create a black-box attack), we also consider training patches that fool an ensemble of detectors.  In this case we replace the objectness loss \eqref{eqn:loss_obj} with the ensemble loss
\begin{align}
\label{eqn:loss_ens}
L_\text{ens}(P) =  \mathbb{E}_{\theta,I} \sum_{i,j}\max \{\mathcal  S_i^{(j)}(\render(I,P))+1, \, 0\}^2,
\end{align}
where $\mathcal S^{(j)}$ denotes the $j$th detector in an ensemble.

\section{Crafting attacks in the digital world}

\paragraph{Datasets and metrics.} We craft attack patches using the COCO dataset,\footnote{We focus on the COCO dataset for its wide diversity of scenes, although we consider the effect of the dataset later.} which contains a total of 123,000 images.
After removing images from the dataset that do not contain people, we then chose a random subset of $10,000$ images for training. We compute average precision (AP) for the category of interest to measure the effectiveness of patches.

\paragraph{Object detectors attacked.} We experiment with both one-stage detectors, \ie, YOLOv2 and YOLOv3, and two-stage detectors, \ie , R50-C4 and R50-FPN, both of which are based on Faster RCNN with a ResNet-50~\cite{he2016deep} backbone. R50-C4 and R50-FPN use different features for region proposal---R50-C4 uses single-resolution features, while R50-FPN uses a multi-scale feature pyramid. For all these detectors, we adopt standard models pre-trained on COCO, in addition to our own models retrained from scratch (models denoted with ``-r'') to test for attack transferability across network weights. Finally, we consider patches crafted using three different ensembles of detectors---\enstwo: YOLOv2 + R50-FPN, \enstwo-r: YOLOv2 + R50-FPN-r, and \ensthree-r: YOLOv2 + YOLOv3 + R50-FPN-r.

\paragraph{Implementation details.} We use PyTorch for implementation, and we start with a random uniform patch of size $3\times250\times150$ (the patch is dynamically re-sized by $\render$ during the forward pass). We use the size since the aspect ratio is similar to that of a body trunk and it has sufficient capacity to reduce objectness scores. We use the Adam optimizer with a learning rate $10^{-3}$, and decay the rate every 100 epochs until 400 is reached. For YOLOv2/v3, images are resized to $640\times640$ for both training and testing. For Faster RCNN detectors, the shortest side of images is $250$\footnote{We found that using a lower resolution produced more effective attacks.} for training, and $800$ for testing.

\subsection{Evaluation of digital attacks}

 \begin{table*}[!t]
     \centering
     \ra{1.0}
     \resizebox{.9\linewidth}{!}{
    \small
     \begin{tabular}{@{}*{16}c@{}}
     \toprule %
      \hspace{2mm}\diagbox[width=2.25cm, height=.7cm]{Patch}{\hbox{Victim}}  & & R50-C4 & R50-C4-r & & R50-FPN & R50-FPN-r & & YOLOv2 & YOLOv2-r & & YOLOv3 & YOLOv3-r \\
     \cmidrule{1-1} \cmidrule{3-4}  \cmidrule{6-7} \cmidrule{9-10} \cmidrule{12-13}
     \textsc{R50-C4}                  &  & \cval{24.5}   & \cval{24.5}     &  & \cval{31.4}    & \cval{31.4}      &  & \cval{37.9}   & \cval{42.6}     &  & \cval{57.6}   & \cval{48.3}     \\
     \textsc{R50-C4}-r                &  & \cval{25.4}   & \cval{23.9}     &  & \cval{30.6}    & \cval{30.2}      &  & \cval{37.7}   & \cval{42.1}     &  & \cval{57.5}   & \cval{47.4}     \\
     \textsc{R50-FPN}                 &  & \cval{20.9}   & \cval{21.1}     &  & \cval{23.5}    & \cval{19.6}      &  & \cval{22.6}   & \cval{12.9}    &  & \cval{40.2}   & \cval{40.3}    \\
     \textsc{R50-FPN}-r               &  & \cval{21.5}   & \cval{21.7}     &  & \cval{25.4}   & \cval{18.8}      &  & \cval{17.6}   & \cval{11.2}     &  & \cval{37.5}  & \cval{36.9}     \\
     \cmidrule{1-1}
     \textsc{Yolov2}                  &  & \cval{21.1}   & \cval{19}       &  & \cval{21.5}    & \cval{21.4}      &  & \cval{10.7}   & \cval{7.5}      &  & \cval{18.1}   & \cval{25.7}     \\
     \textsc{Yolov3}                  &  & \cval{28.3}   & \cval{28.9}     &  &\cval{ 31.5}    & \cval{27.2}      &  & \cval{20}     & \cval{15.9}     &  & \cval{17.8}   & \cval{36.1}     \\
     \cmidrule{1-1} 
     \ft            &  & \cval{25.6}   & \cval{23.9}     &  & \cval{24.2}    & \cval{24.4}      &  & \cval{18.9}   & \cval{16.4}     &  & \cval{31.6}   & \cval{28.2}     \\
     \enstwo               &  & \cval{20}     & \cval{20.3}     &  & \cval{23.2}    & \cval{19.3}      &  & \cval{17.5}  & \cval{11.3}   &  & \cval{39}    & \cval{38.8}     \\
     \enstwo-r             &  & \cval{19.7}   & \cval{20.2}     &  & \cval{23.3 }   & \cval{16.8}      &  & \cval{14.9}   & \cval{9.7 }     &  & \cval{36.3}   & \cval{34.1}     \\
     \ensthree-r             &  & \cval{21.1}   & \cval{21.4}    &  & \cval{24.2}    & \cval{17.4 }     &  & \cval{13.4}   & \cval{9.0 }       &  & \cval{29.8}   & \cval{33.6 }    \\
     \cmidrule{1-1} 
     \textsc{Seurat}                  &  & \cval{47.9}   & \cval{52}       &  & \cval{51.6}    & \cval{52.5}      &  & \cval{43.4}   & \cval{39.5}     &  & \cval{62.6}   & \cval{57.1}     \\
     \textsc{Random}                  &  & \cval{53}     & \cval{58.2}     &  & \cval{59.8}    & \cval{59.7}      &  & \cval{52}     & \cval{52.5}     &  & \cval{70 }    & \cval{63.5}     \\
     \textsc{Grey}                    &  & \cval{45.9}   & \cval{49.6}     &  & \cval{50}      & \cval{50.8}      &  & \cval{48}     & \cval{47.1}     &  & \cval{65.6}   & \cval{57.5}     \\
     \textsc{Grey++}                  &  & \cval{46.5}   & \cval{49.8}     &  & \cval{51.4}    & \cval{52.7 }     &  & \cval{48.5}   & \cval{49.4}     &  & \cval{64.8}   & \cval{58.6}     \\
     \cmidrule{1-1} 
     \textsc{Clean}                   &  & \cval{78.7}   & \cval{78.7}     &  & \cval{82.2}    & \cval{82.1}      &  & \cval{63.6}   & \cval{62.7}     &  & \cval{81.6}   & \cval{74.5}     \\ \hline
     \end{tabular}}
     \caption{\textbf{Impact of different patches on various detectors, measured using average precision (AP).} The left axis lists patches created by different methods, and the top axis lists different victim detectors.  Here, ``r'' denotes retrained weights instead of pretrained weights downloaded from model zoos.  
     }
     \label{tbl:digital_white}
 \end{table*}

We begin by evaluating patches in digital simulated settings: we consider white-box attacks (detector weights are used for patch learning) and black-box attacks (patch is crafted on a surrogate model and tested on a victim model with different parameters). 

\begin{figure*}[h]
    \centering
    \resizebox{0.8\linewidth}{!}{
    \begin{subfigure}[b]{0.12\linewidth}
    \includegraphics[width=1.0\linewidth]{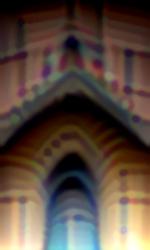}
    \caption{C4}
    \end{subfigure}
    \begin{subfigure}[b]{0.12\linewidth}
    \includegraphics[width=1.0\linewidth]{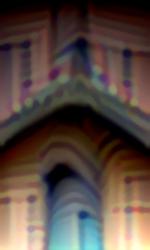}
    \caption{C4-r}
    \end{subfigure}
    \begin{subfigure}[b]{0.12\linewidth}
        \includegraphics[width=1.0\linewidth]{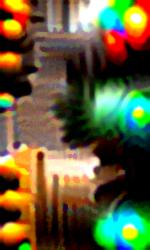}
        \caption{FPN}
    \end{subfigure}
    \begin{subfigure}[b]{0.12\linewidth}
        \includegraphics[width=1.0\linewidth]{gallery/patches/R50-FPN}
        \caption{FPN-r}
    \end{subfigure}
    \begin{subfigure}[b]{0.12\linewidth}
        \includegraphics[width=1.0\linewidth]{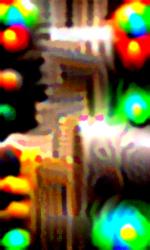}
        \caption{Ens2}
    \end{subfigure}
    \begin{subfigure}[b]{0.12\linewidth}
        \includegraphics[width=1.0\linewidth]{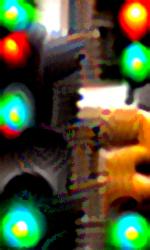}
        \caption{Ens2-r}
    \end{subfigure}
    \begin{subfigure}[b]{0.12\linewidth}
        \includegraphics[width=1.0\linewidth]{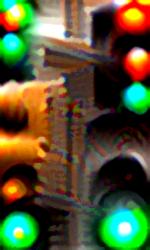}
        \caption{Ens3-r}     
    \end{subfigure}}
    \\
    \resizebox{0.8\linewidth}{!}{
    \begin{subfigure}[b]{0.12\linewidth}
        \includegraphics[width=1.0\linewidth]{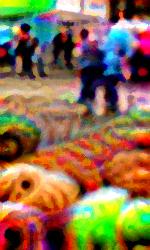}
        \caption{Fted}
    \end{subfigure}
        \begin{subfigure}[b]{0.12\linewidth}
        \includegraphics[width=1.0\linewidth]{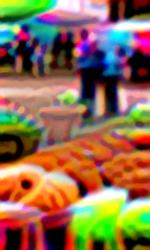}
        \caption{Yolov2}
        \label{fig:patch_yolov2}
    \end{subfigure}
    \begin{subfigure}[b]{0.12\linewidth}
        \includegraphics[width=1.0\linewidth]{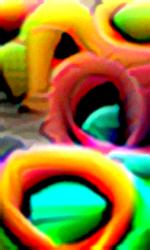}
        \caption{Yolov3}
    \end{subfigure}
       \begin{subfigure}[b]{0.12\linewidth}
        \includegraphics[width=1.0\linewidth]{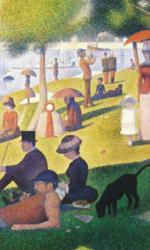}
        \caption{Seurat}
        \label{fig:patch_seurat}
    \end{subfigure}
    \begin{subfigure}[b]{0.12\linewidth}
        \includegraphics[width=1.0\linewidth]{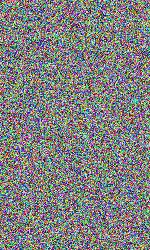}
        \caption{Rand}
    \end{subfigure}
     \begin{subfigure}[b]{0.12\linewidth}
        \includegraphics[width=1.0\linewidth]{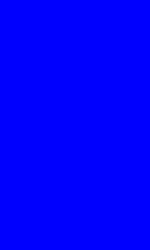}
        \caption{G++}
    \end{subfigure}
    \begin{subfigure}[b]{0.12\linewidth}
        \includegraphics[width=1.0\linewidth]{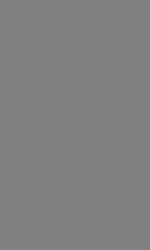}
        \caption{Grey}
        \label{fig:patches_grey}
    \end{subfigure}}
    \caption{\textbf{Adversarial patches}, and comparisons with control patches. Here, (a)-(d) are based on R50, and G++ denotes Grey++.   }
    \label{fig:patches}
\end{figure*}

\paragraph{Effectiveness of learned patches for white-box attack.} We optimize patches using the aforementioned detectors, and denote the learned patch with the corresponding model it is trained on. We further compare with the following alternative patches:
\begin{enumerate*}[label=(\arabic*)]
    \item \textsc{Fted}, a learned \yt patch that is further fine-tuned on a R50-FPN model;
    \item \textsc{Seurat}, a crop from the famous paining ``A Sunday Afternoon on the Island of La Grande Jatte'' by Georges Seurat, which is visually similar to the top-performing \yt patch with objects like persons, umbrellas \etc (see Fig.~\ref{fig:patch_seurat});
    \item \textsc{Grey}, a grey patch;
    \item \textsc{Grey++}, the most powerful RGB value for attacking Yolov2 using  COCO;
    \item \textsc{Random}, a randomly initialized patch;
    \item \textsc{Clean}, which corresponds to the oracle performance of detectors when patches are not applied.
\end{enumerate*}

\begin{figure}
\centering
\resizebox{0.9\linewidth}{!}{
\begin{minipage}[t]{.48\textwidth}
  \centering
  \includegraphics[width=1.0\linewidth]{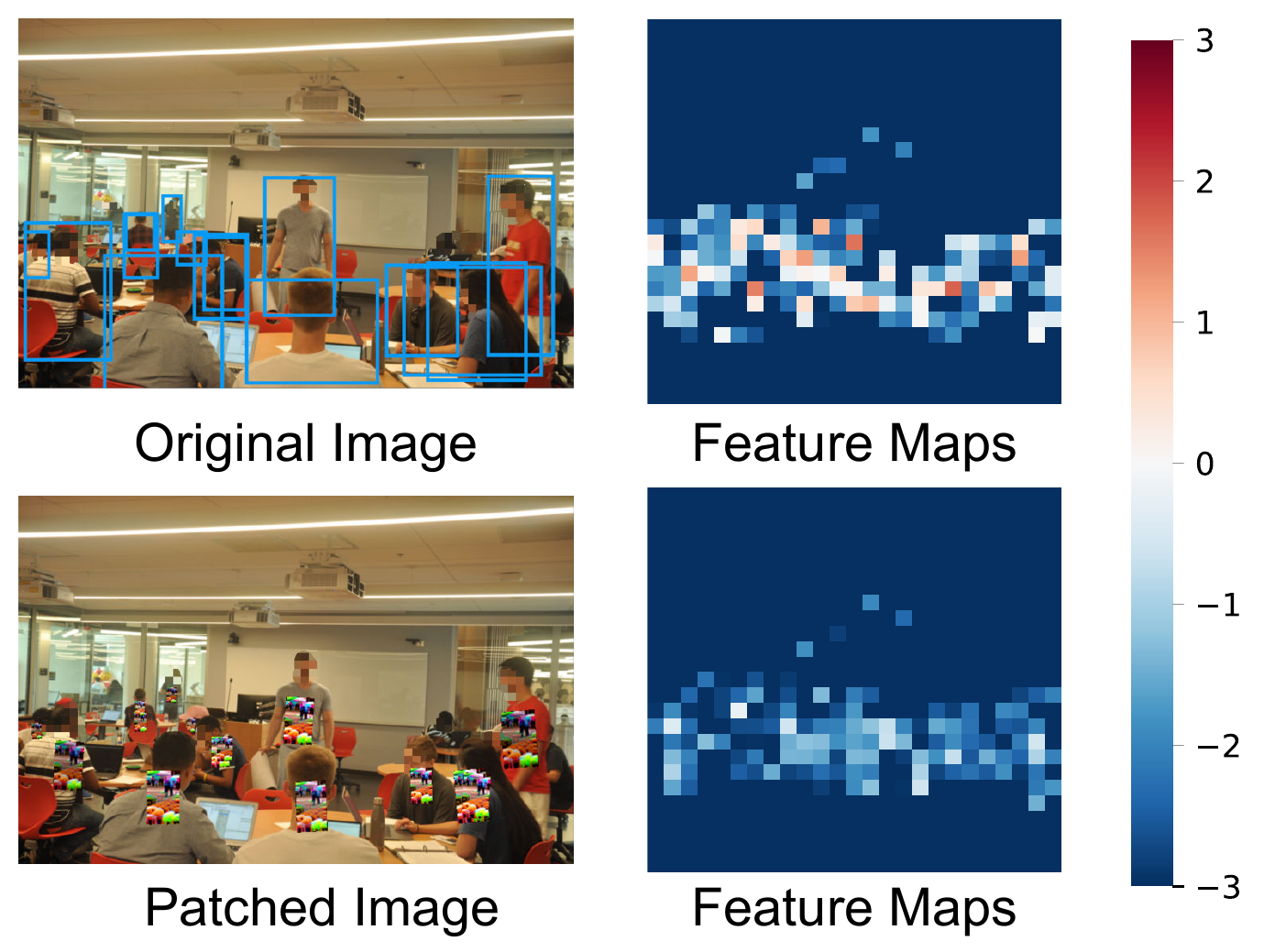}
\end{minipage}
~
\begin{minipage}[t]{0.48\textwidth}
  \centering
  \includegraphics[width=1.0\linewidth]{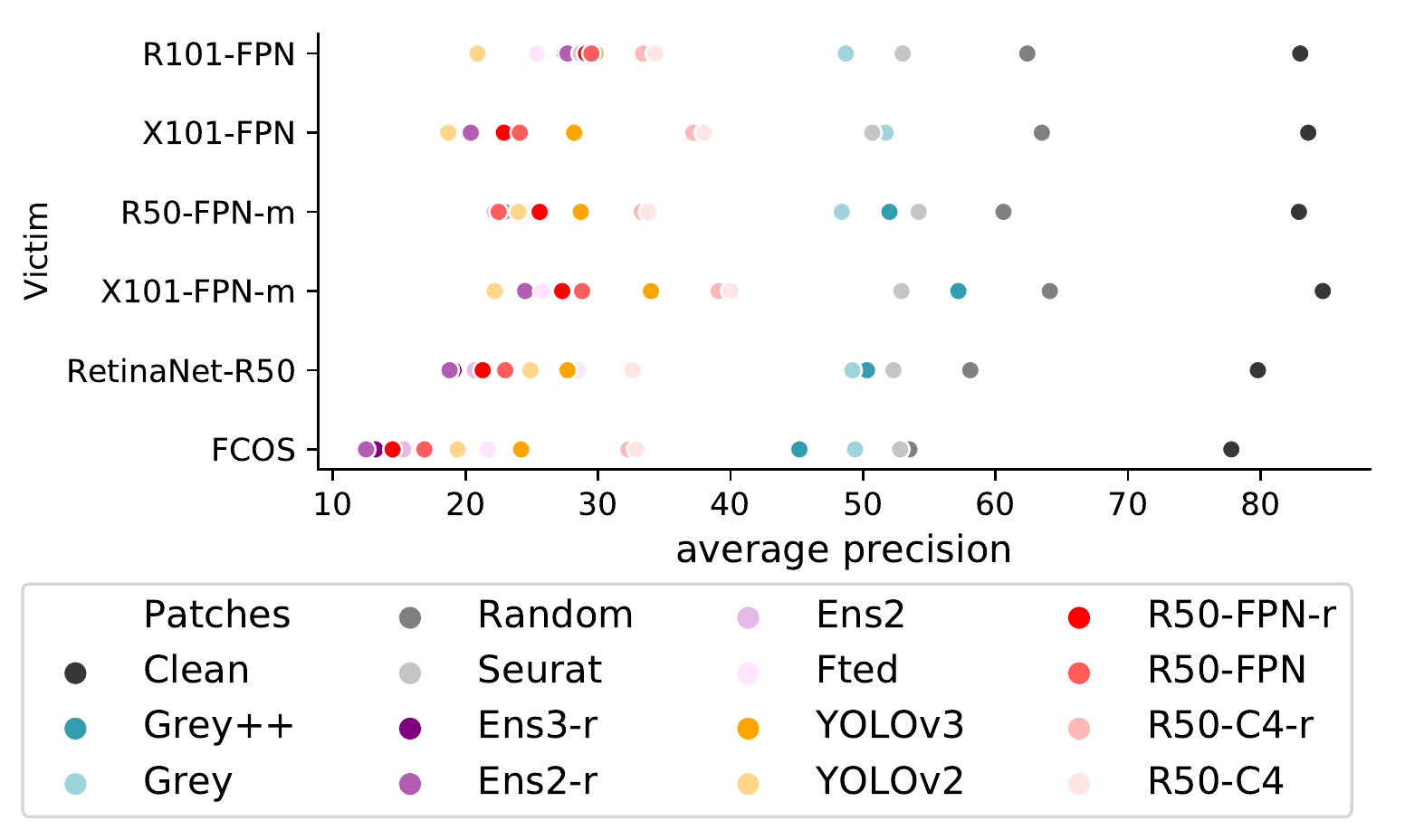}
\end{minipage}}
\\
\resizebox{1.0\linewidth}{!}{
\begin{minipage}[t]{0.48\textwidth}
  \centering
  \caption{\textbf{Images and their feature maps},  w. and w/o patches, using YOLOv2. Each pixel in the feature map represents an objectness score.}
\label{fig:feat_map}
\end{minipage}
~~
\begin{minipage}[t]{0.48\textwidth}
  \centering
  \caption{\textbf{Performance of different patches}, when tested on detectors with different backbones.}
\label{fig:transfer_to_det}
\end{minipage}}
\end{figure}

Adversarial and control patches are shown in Fig.~\ref{fig:patches}, and results are summarized in Table~\ref{tbl:digital_white}. We observe that all adversarially learned patches are highly effective in digital simulations, where the AP of all detectors degrades by at least 29\%, going as low as 7.5\% AP when the \yt patch is tested on the retrained \yt model (YOLOv2-r). All patches transferred well to the corresponding retrained models. In addition, the ensemble patches perform better compared to Faster RCNN patches but are worse than YOLO patches. It is also interesting to see that \textsc{Yolo} patches can be effectively transferred to Faster RCNN models, while Faster RCNN patches are not very effective at attacking YOLO models. Although the \textsc{Seurat} patch is visually similar to the learned \yt patch (cf. Fig.~\ref{fig:patch_yolov2} and Fig.~\ref{fig:patch_seurat}), it does not consistently perform better than \textsc{Grey}.  We visualize the impact of the patch in Fig~\ref{fig:feat_map}, which shows objectness maps from the YOLOv2 model with and without patches. We see that when patches are applied to persons, the corresponding pixels in the feature maps are suppressed. 

\paragraph{Transferability across backbones.} We also investigate whether the learned patches transfer to detectors with different backbones. We evaluate the patches on the following detectors:
\begin{enumerate*}[label=(\arabic*)]
    \item \textsc{R101-FPN}, which uses ResNet-101 with FPN as its backbone;
    \item \textsc{X101-FPN}, replaces the \textsc{R101-FPN} with ResNeXt-101~\cite{xie2017aggregated};
    \item \textsc{R50-FPN-m}, a Mask RCNN model~\cite{he2017mask} based on R50-FPN;
    \item \textsc{X101-FPN-m}, a Mask RCNN model based on X101-FPN;
    \item \textsc{RetinaNet-R50}, a RetinaNet~\cite{lin2017focal} with a backbone of ResNet-50;
    \item \textsc{FCOS}, a recent anchor-free framework~\cite{tian2019fcos} based on R50-FPN.
\end{enumerate*} 
The results are shown in Fig~\ref{fig:transfer_to_det}. We observe that all these learned patches can significantly degrade the performance of the person category even using models that they have not been trained on.

\begin{table*}[h]
\centering
\ra{1.1}
\resizebox{1.0\linewidth}{!}{
    \begin{tabular}{@{}*{23}c@{}}
    \toprule
    \hspace{2mm}\diagbox[width=2.25cm, height=.7cm]{Patch}{\hbox{Class}} && aero & bike & bird & boat & bottle & bus  & car  & cat  & chair & cow  & table & dog  & horse & mbike & person & plant & sheep & sofa & train & tv   \\ 
    \cmidrule{1-1} \cmidrule{3-22}
    \textsc{Person} && \cval{2.0 } & \cval{14.6} & \cval{1.0 } & \cval{1.8 } & \cval{2.7 } & \cval{13.5} & \cval{10.7} & \cval{2.3 } & \cval{0.1 } & \cval{2.4 } & \cval{6.4 } & \cval{2.3 } & \cval{8.3 } & \cval{12.3} & \cval{5.5 } & \cval{0.3 } & \cval{2.2 } & \cval{1.3 } & \cval{3.8 } & \cval{12.4} \\
    \textsc{Horse}  && \cval{5.0 } & \cval{31.9} & \cval{4.7 } & \cval{4.1 } & \cval{2.5 } & \cval{26.4} & \cval{17.6} & \cval{10.6} & \cval{2.3 } & \cval{26.0} & \cval{24.7} & \cval{9.5 } & \cval{27.9} & \cval{26.6} & \cval{16.0} & \cval{7.6 } & \cval{12.4} & \cval{13.4} & \cval{13.2} & \cval{35.3} \\
    \textsc{Bus}    && \cval{3.1 } & \cval{30.6} & \cval{8.5 } & \cval{4.4 } & \cval{1.9 } & \cval{18.4} & \cval{15.6} & \cval{7.8 } & \cval{2.7 } & \cval{25.7} & \cval{39.8} & \cval{5.3 } & \cval{20.8} & \cval{20.7} & \cval{16.0} & \cval{8.9 } & \cval{12.3} & \cval{9.5 } & \cval{9.3 } & \cval{29.5} \\
    \cmidrule{1-1}
    \textsc{Grey}   && \cval{3.0 } & \cval{19.0} & \cval{6.4 } & \cval{14.6} & \cval{8.5 } & \cval{26.9} & \cval{19.6} & \cval{9.9 } & \cval{9.8 } & \cval{28.6} & \cval{24.4} & \cval{7.4 } & \cval{22.7} & \cval{15.9} & \cval{35.8} & \cval{6.1 } & \cval{18.7} & \cval{8.7 } & \cval{11.4} & \cval{61.8} \\ 
    \cmidrule{1-1}
    \textsc{Clean}  && \cval{77.5} & \cval{82.2} & \cval{76.3} & \cval{63.6} & \cval{64.5} & \cval{82.9} & \cval{86.5} & \cval{83.0} & \cval{57.2} & \cval{83.3} & \cval{66.2} & \cval{84.9} & \cval{84.5} & \cval{81.4} & \cval{83.3} & \cval{48.0} & \cval{76.7} & \cval{70.1} & \cval{80.1} & \cval{75.4} \\
    \bottomrule
    \end{tabular}}
    \caption{\textbf{Transferability of patches across classes from VOC}, measured with average precision (AP).}
    \label{tbl:attack_classes}
    \end{table*}

\begin{figure}[t]
\centering
\resizebox{0.8\linewidth}{!}{
\includegraphics[width=0.75\linewidth]{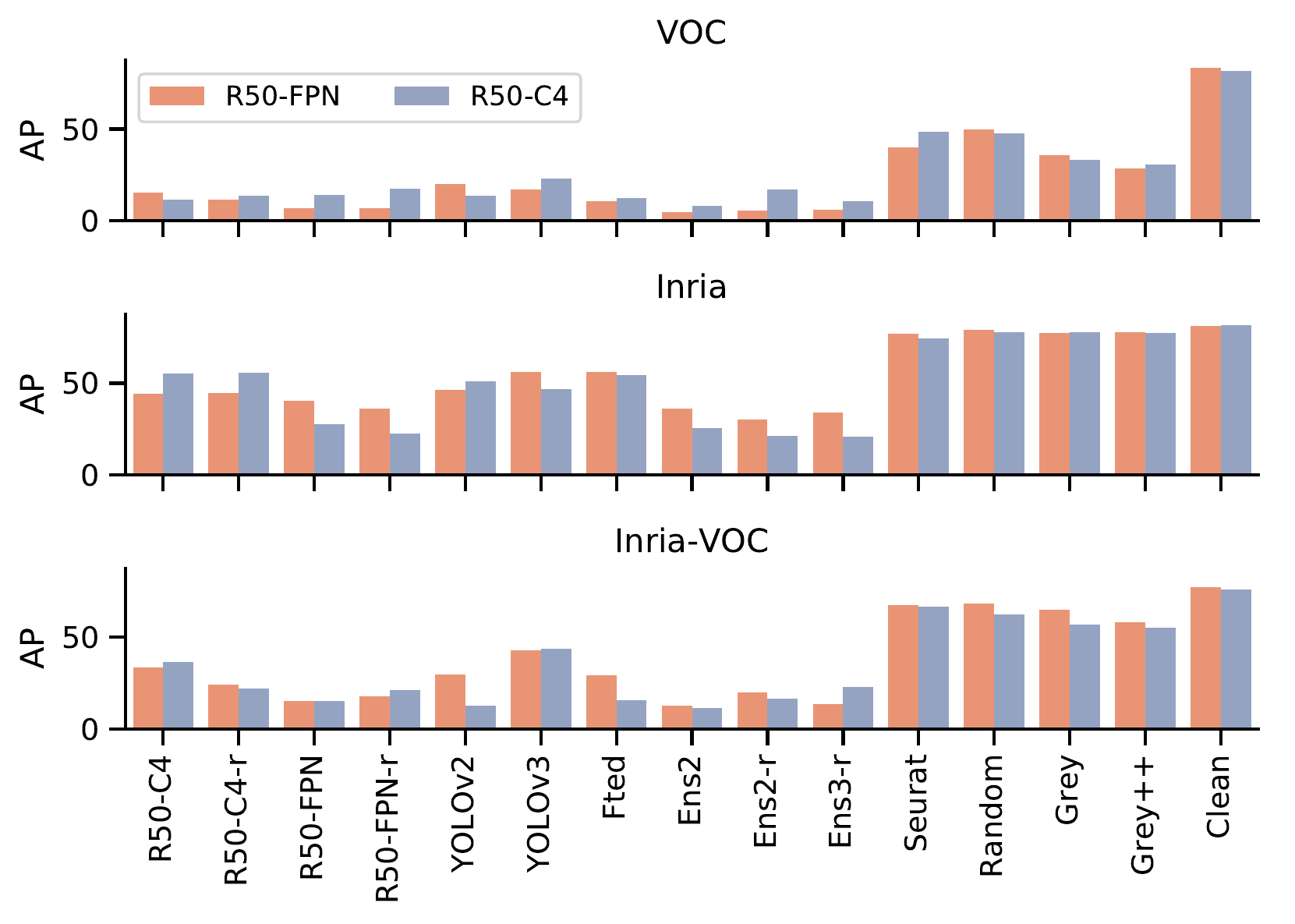}}
\caption{\textbf{Results of different patches, trained on COCO, tested on the person category of different datasets}. Top two panels: COCO patches tested on VOC and Inria, respectively, using backbones learned on COCO; The bottom panel: COCO patches tested on Inria with backbones trained on VOC.}
\label{fig:datasets_transfer}
\end{figure}

\paragraph{Transferability across datasets.} We further demonstrate the transferability of patches learned on COCO to other datasets including Pascal VOC 2007~\cite{everingham2015pascal} and the Inria person dataset~\cite{DBLP:conf/cvpr/DalalT05}. We evaluate the patches on the person category using R50-FPN and R50-C4, and the results are presented in Fig.~\ref{fig:datasets_transfer}. The top two panels correspond to results of different patches when evaluated on VOC and Inria, respectively, with both the patches and the models trained on COCO; the bottom panel shows the APs of these patches when applied to Inria images using models trained on VOC. We see that ensemble patches offer the most effective attacks, degrading the AP of the person class by large margins. This confirms that these patches can transfer not only to different datasets but also backbones trained with different data distributions. From the bottom two panels, we can see that weights learned on COCO are more robust than on VOC.

\paragraph{Transferability across classes.} We find that patches effectively suppress multiple classes, even when they are trained to suppress only one. In addition to the ``person'' patch, we also train patches on  ``bus'' and ``horse,'' and then evaluate these patches on all 20 categories in VOC.\!\footnote{We observe similar trends on COCO.} Table~\ref{tbl:attack_classes} summarizes the results. We can see that the ``person'' patch transfers to almost all categories, possibly because they co-occur with most classes. We also compare with the \textsc{Grey} patch to rule out the possibility that the performance drops are due to occlusion.

\section{Physical world attacks}
\label{sec:exp_physical}
\begin{figure}[h]
\centering
\resizebox{0.9\linewidth}{!}{ 
    \begin{minipage}{1\linewidth}
    \centering
    \begin{subfigure}[b]{0.3\linewidth}
    \includegraphics[width=1\linewidth]{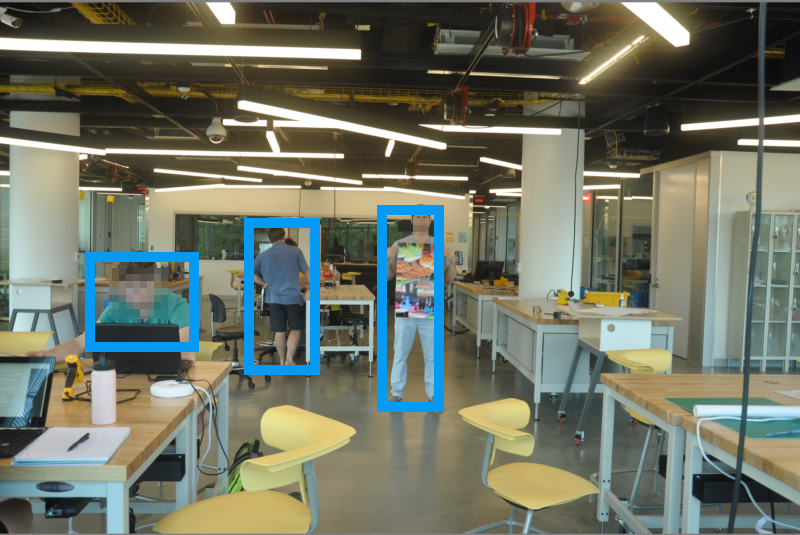}
    \caption{Failure-p}
    \end{subfigure}
    \begin{subfigure}[b]{0.3\linewidth}
        \includegraphics[width=1\linewidth]{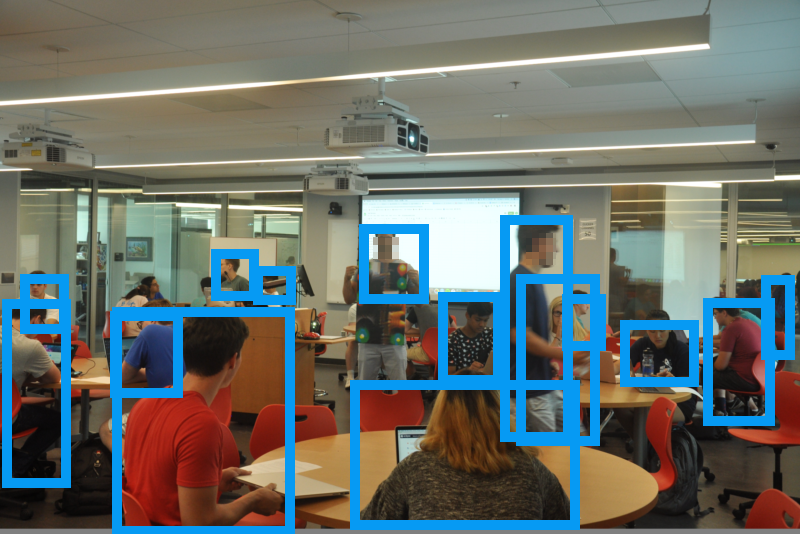}
            \caption{Partial-p}
        \end{subfigure}
        \begin{subfigure}[b]{0.3\linewidth}
            \includegraphics[width=1\linewidth]{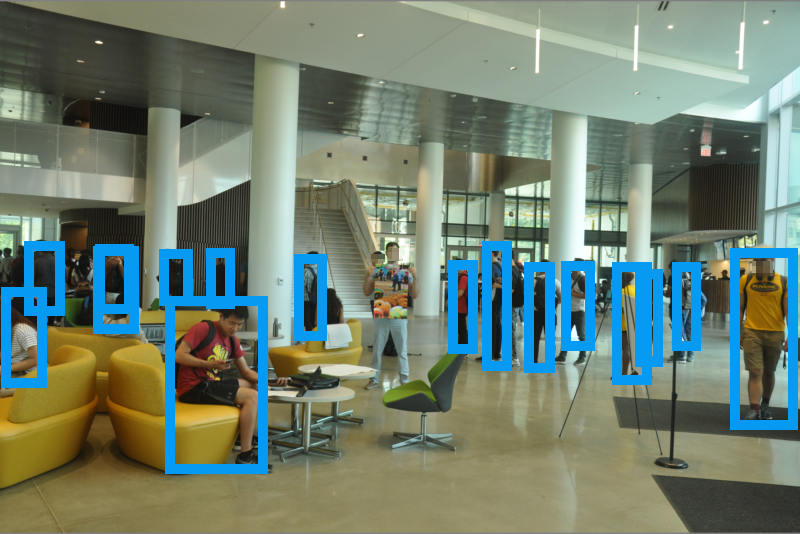}
                    \caption{Success-p}
            \end{subfigure}
            \\
    \begin{subfigure}[b]{0.3\linewidth}
        \includegraphics[width=1\linewidth]{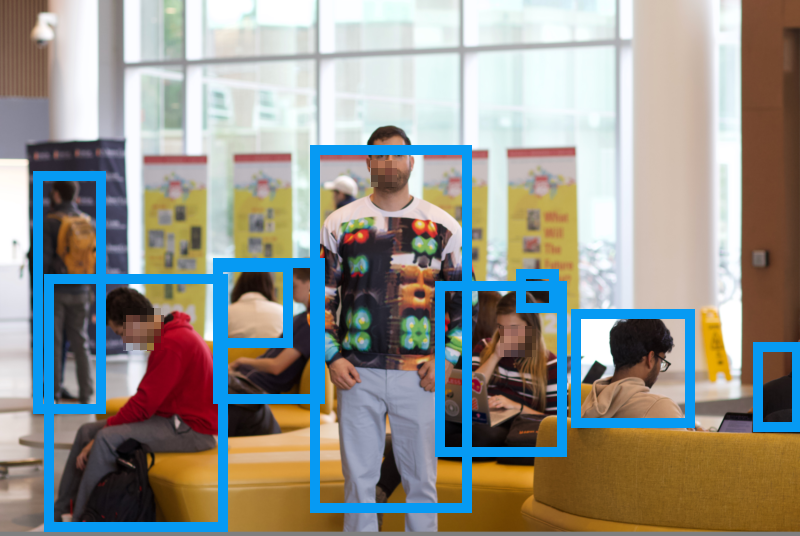}
            \caption{Failure-c}
        \end{subfigure}
        \begin{subfigure}[b]{0.3\linewidth}
            \includegraphics[width=1\linewidth]{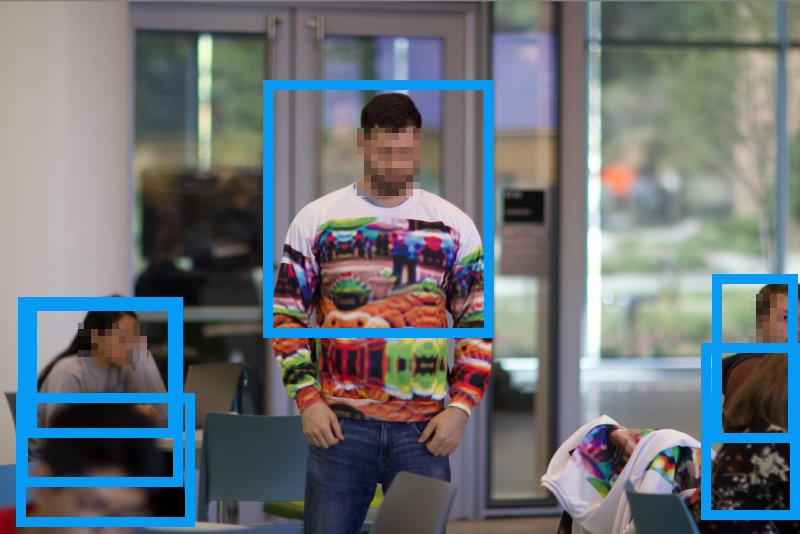}
                    \caption{Partial-c}
            \end{subfigure}
            \begin{subfigure}[b]{0.3\linewidth}
                \includegraphics[width=1\linewidth]{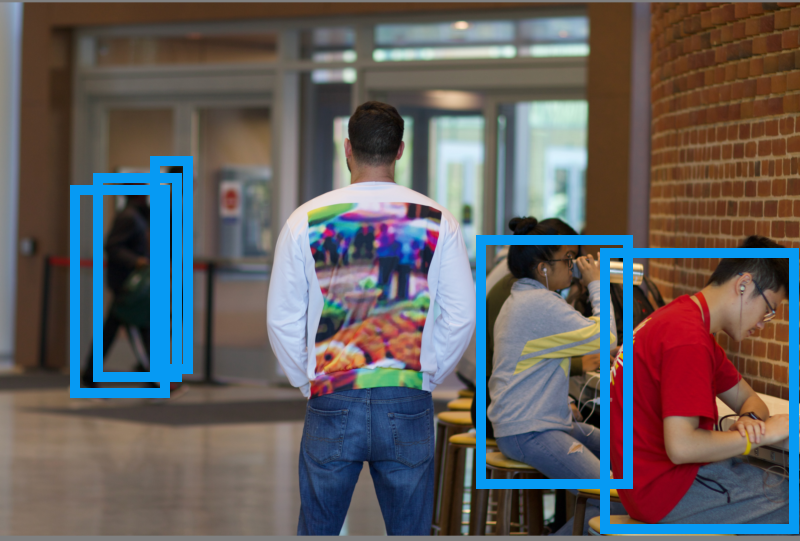}
                            \caption{Success-c}
                \end{subfigure}
    \end{minipage}}
                \\
   \resizebox{1.0\linewidth}{!}{
    \begin{minipage}{1.0\linewidth}
    \centering
    \caption{\textbf{Examples of attack failure, partial success, and full success}, using posters  (top) and shirts (bottom).}
    \label{fig:example}
    \end{minipage}}
\end{figure}

We now discuss the results of physical world attacks with printed posters.  In addition to the standard average precision,\!\footnote{We only consider the person with adversarially patterns to calculate AP by eliminating boxes without any overlapping with the GT box.} we also quantify the performance of attacks with ``success rates,''   which we define as
\begin{enumerate*}[label=(\arabic*)]
    \item a \textsc{Success} attack: when there is no bounding box predicted for the person with adversarial patterns;
    \item a \textsc{Partial success} attack: when there is a bounding box covering less than $50\%$ of a person;
    \item a \textsc{Failure} attack: when the person is successfully detected.
\end{enumerate*}
Examples of detections in each category are shown in Fig.~\ref{fig:example}.
To compute these scores, we use cutoff zero for YOLOv2, and we tune the threshold of other detectors to achieve the best F-1 score on the COCO minival set. 

\subsection{Printed posters} We printed posters and took photos at 15 different locations using 10 different patches. At each location, we took four photos for each printed patch corresponding to two distances from the camera and two heights where the patch is held.  We also took photos without printed posters as controls (\textsc{Control}). In total, we collected 630 photos (see the top row of Figure~\ref{fig:example} for examples). We use four patches that perform well digitally (\ie, \yt, \enstwo, \ensthree, \textsc{Fted}), and three baseline patches (\textsc{Seurat} patch, \textsc{Flip} patch, \textsc{White}). 

To better understand the impact of the training process on attack power, we also consider several variants of the \yt patch (the best digital performer). 
To assess whether the learned patterns are ``truly'' adversarial, or whether any qualitatively similar pattern would work, we add the \textsc{Flip} patch, which is the \yt patch held upside-down.   We compare to a \textsc{TPS} patch, which uses thin plate spline transformations to potentially enhance robustness to warping in real objects.  We consider a \yt-noaug patch, which is trained to attack the YOLOv2 model without any augmentations/transformations beyond resizing. To observe the effect of the dataset, we add the \yt-Inria patch, which is trained on the Inria dataset as opposed to COCO.

\begin{figure*}[t!]
    \centering
    \begin{subfigure}[b]{0.45\linewidth}
    \includegraphics[width=1.0\linewidth]{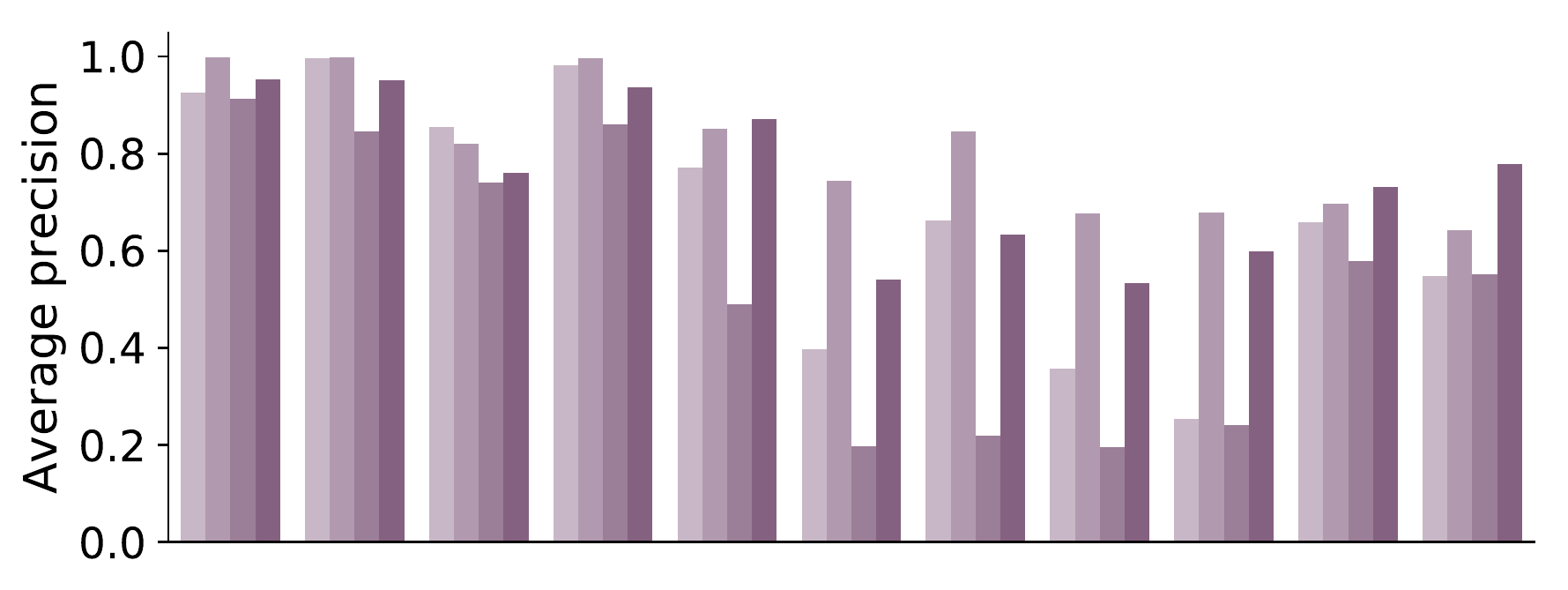}
    \caption{AP of posters}
    \label{fig:ap_poster}
    \end{subfigure}
    \begin{subfigure}[b]{0.45\linewidth}
    \includegraphics[width=1.0\linewidth]{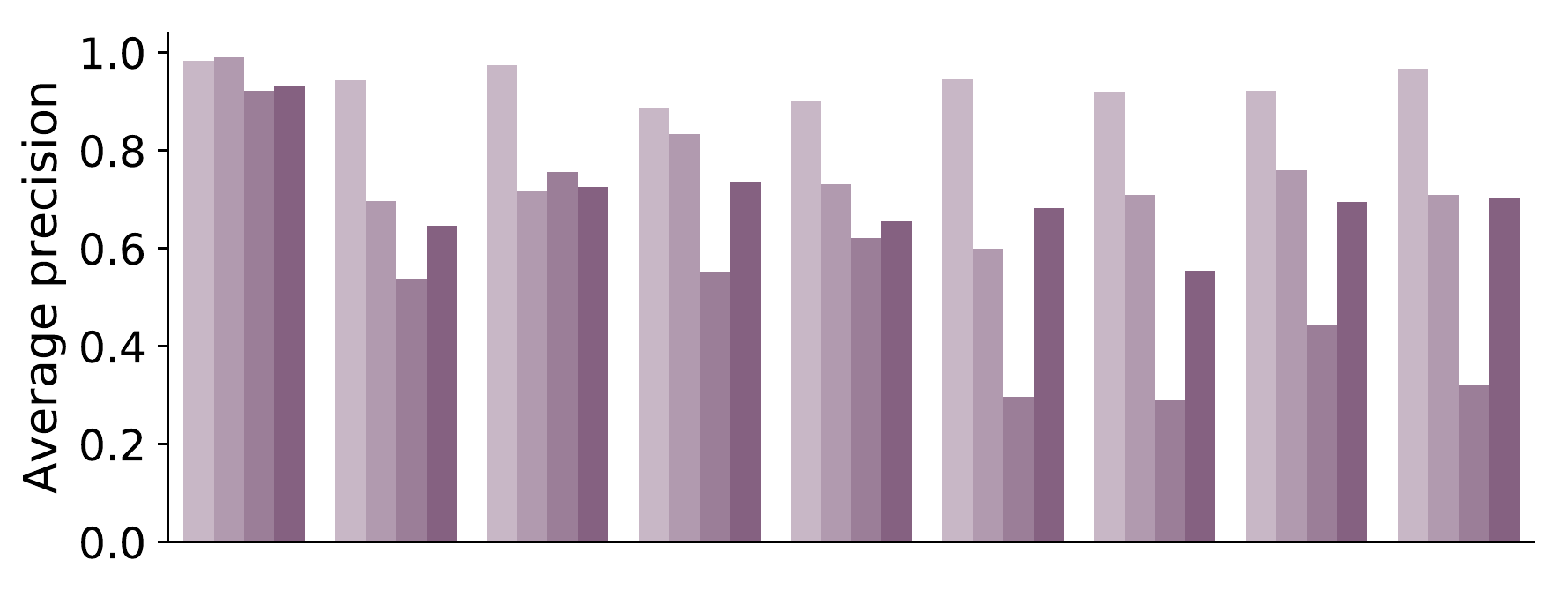}
    \caption{AP of clothes}
    \label{fig:ap_clothes}
    \end{subfigure}
    \\
    \begin{subfigure}[t]{0.45\linewidth}
    \includegraphics[width=1.0\linewidth]{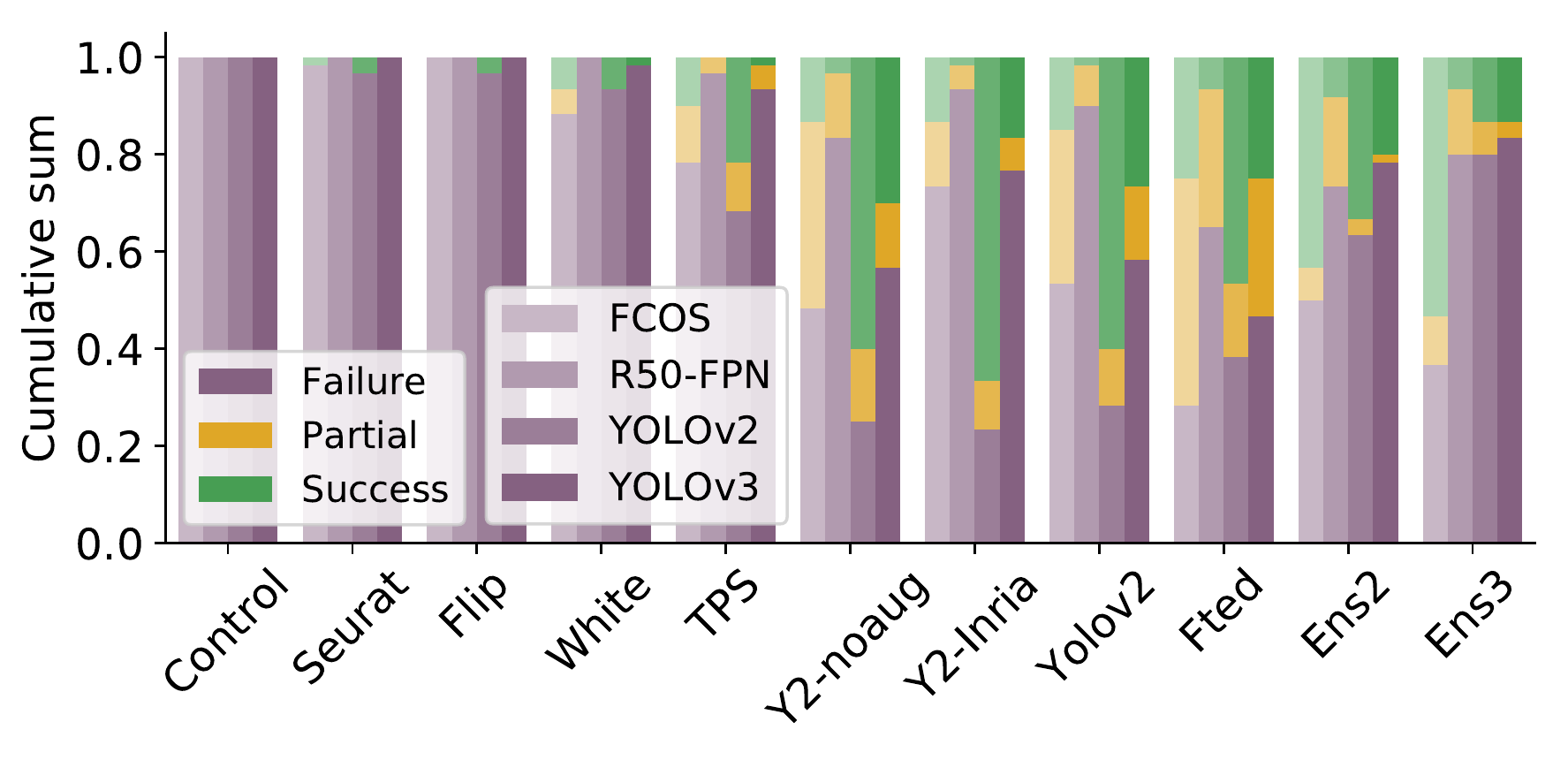}
    \caption{Success rates of posters}
    \label{fig:success_poster}
    \end{subfigure}
    \begin{subfigure}[t]{0.45\linewidth}
        \includegraphics[width=1.0\linewidth]{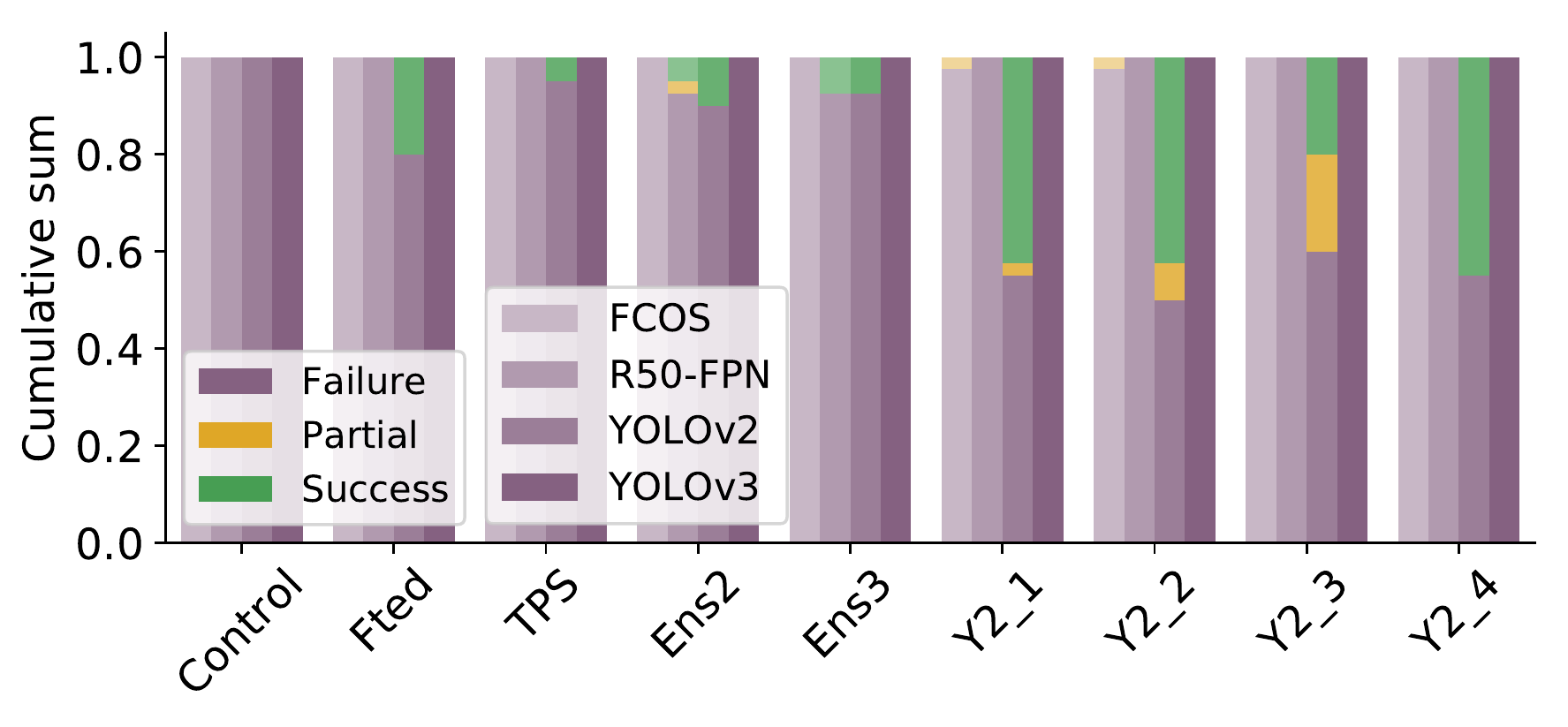}
        \caption{Success rates of clothes}
        \label{fig:success_clothes}
    \end{subfigure}
    \caption{\textbf{AP and success rates for physical attacks}. Top: AP of different printed posters (left) and clothes (right). Lower is better. Bottom: success rates of different printed posters (left) and clothes (right). Y2 denotes \yt.}
    \label{fig:physical_attacks}
\end{figure*}

\paragraph{Poster results.} Figure~\ref{fig:ap_poster} and~\ref{fig:success_poster} summarize the results. We can see that compared to baseline patches, Adversariallyy optimized patches successfully degrade the performance of detectors measured by both AP and success rates.  
The \yt patch achieves the best performance measured by AP among all patches. R50-FPN is the most robust model with slight degradation when patches are applied.  FCOS is the most vulnerable network; it fell to the \yt patch even though we never trained on an anchor-free detector, let alone FCOS. This may be because anchor-free models predict the ``center-ness'' of pixels for bounding boxes, and the performance drops when center pixels of persons are occluded by printed posters. Interestingly though, simply using baseline patches for occlusion fails to deceive FCOS. 

Beyond the choice of detector, several other training factors impact performance.  Surprisingly, the \textsc{TPS} patch is worse than \yt, and we believe this results from the fact that adding such complicated transformation makes optimization more difficult during training. It is also surprising to see that the \yt-Inria patch offers impressive success rates on \yt, but it does not transfer as well to other detectors. Not surprisingly, the \yt patch outperforms the \yt-noaug in terms of AP, however these gains shrink when measured in terms of success rates.

We included the \textsc{Flip} patch to evaluate whether patches are generic, \ie, any texture with similar shapes and scales would defeat the detector, or whether they are  ``truly adversarial.''  The poor performance of the \textsc{Flip}  patch seems to indicate that the learned patches are exploiting specialized behaviors learned by the detector, rather than a generic weakness of the detector model. From the left column of Figure~\ref{fig:physical_attacks} and Table~\ref{tbl:digital_white}, we see that performance in digital simulations correlates well with physical world performance. However, we observe that patches lose effectiveness when transferring from the digital world into the physical world, demonstrating that physical world attacks are more challenging. 

\subsection{Paper dolls}  We found that a useful technique for crafting physical attacks was to make ``paper dolls''---printouts of images that we could dress up with different patches at different scales.  This facilitates  quick experiments with physical world effects and camera distortions without the time and expense of fabricating textiles. We use paper dolls to gain insights into why physical attacks are not as effective as digital attacks. 
The reasons might be three-fold: 
\begin{enumerate*}[label=(\arabic*)]
    \item Pixelation at the detector and compression algorithms incur subtle changes;
    \item the rendering artifacts around patch borders assist digital attacks;
    \item there exists differences in appearance and texture between large-format digital patches and the digital patches.
\end{enumerate*} 

\begin{figure}[h]
    \begin{minipage}{1.0\linewidth}   
    \centering 
    \includegraphics[width=0.6\linewidth]{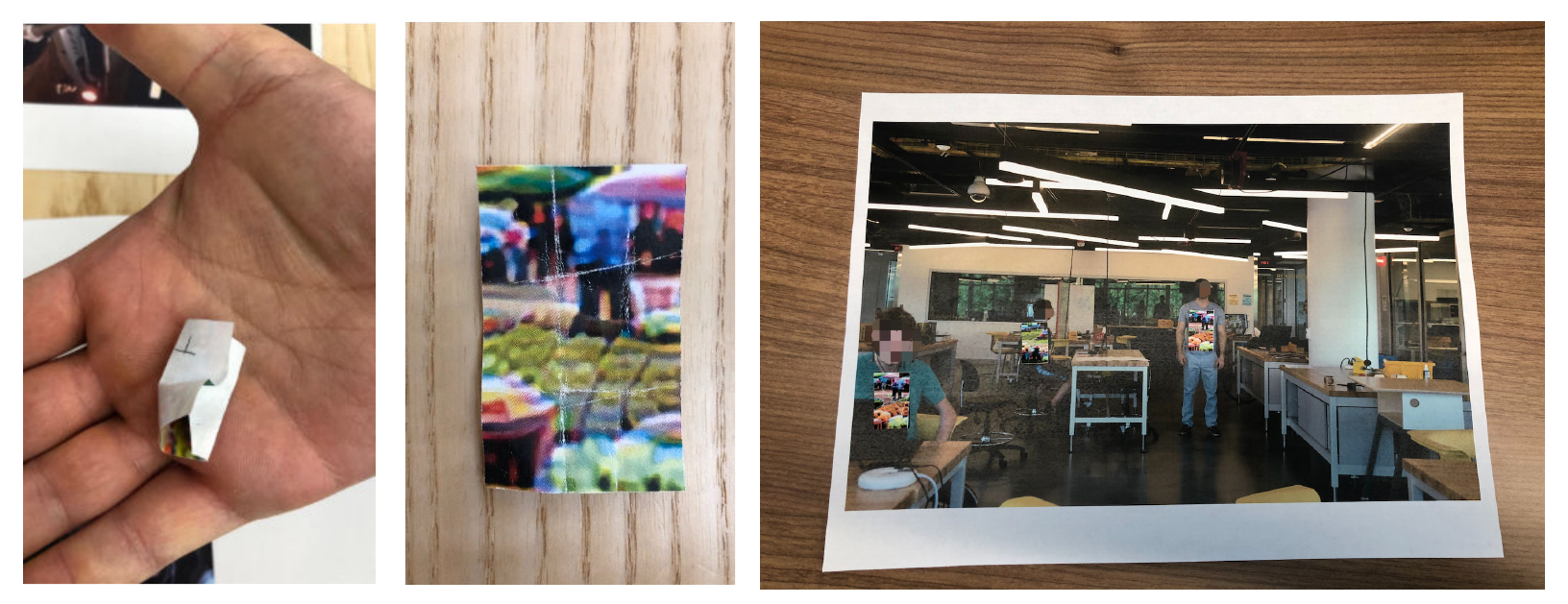}
    \caption{\textbf{Paper dolls} are made by dressing up printed images with paper patches.  We use dolls to observe the effects of camera distortions, and ``scrumpled'' patches to test against physical deformations that are not easily simulated.}
    \label{fig:paper_doll}
    \end{minipage}
\end{figure}

In our paper doll study, we print out patches and photos separately. We then overlay patches onto objects and photograph them. We used the first 20 images from the COCO minival set. We use the same patches from the poster experiment, we also compare with ``scrumpled'' versions of \yt, \ie, \yt-s1 and \yt-s2, to test for robustness to physical deformation, where ``-s1'' and ``-s2'' denote the level of crumpling (``s1'' $<$ ``s2'', see Fig.~\ref{fig:paper_doll}).

\begin{figure}
\begin{minipage}[b]{0.48\linewidth}
    \centering 
    \includegraphics[width=0.9\linewidth]{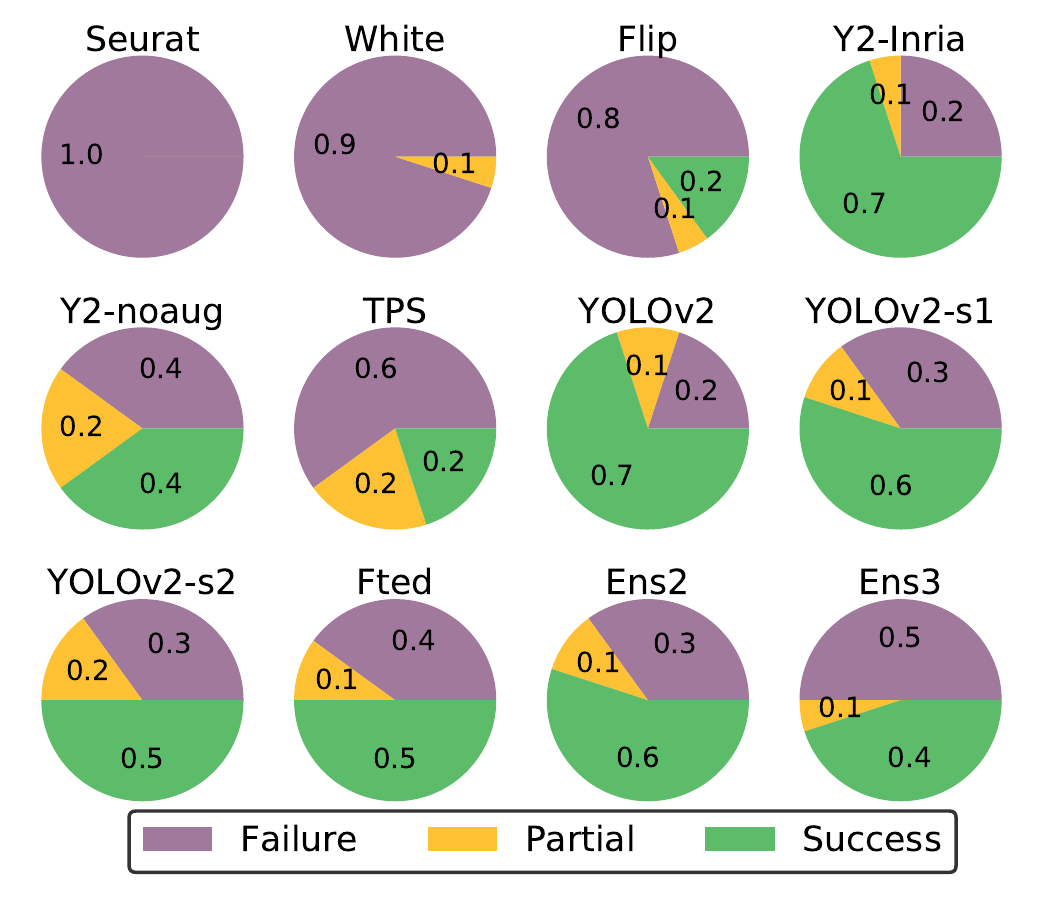}
    \caption{\textbf{Effectiveness of different patches on paper dolls}.}
    \label{fig:pie_dolls}
\end{minipage}
~
\begin{minipage}[b]{0.48\linewidth}
    \centering
    \begin{subfigure}[t]{0.23\linewidth}
    \includegraphics[width=1.0\linewidth]{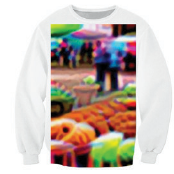}
    \caption{Y2-1}
    \end{subfigure}
    \begin{subfigure}[t]{0.23\linewidth}
    \includegraphics[width=1.0\linewidth]{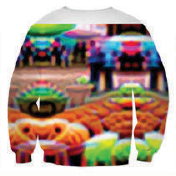}
    \caption{Y2-2}
    \end{subfigure}
    \begin{subfigure}[t]{0.23\linewidth}
        \includegraphics[width=1.0\linewidth]{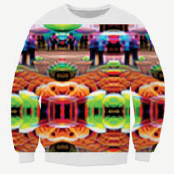}
        \caption{Y2-3}
    \end{subfigure}
    \begin{subfigure}[t]{0.23\linewidth}
        \includegraphics[width=1.0\linewidth]{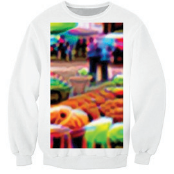}
        \caption{Y2-4}
    \end{subfigure}
    \\
    \begin{subfigure}[t]{0.23\linewidth}
        \includegraphics[width=1.0\linewidth]{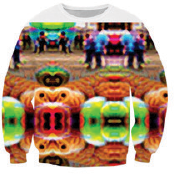}
        \caption{\textsc{fted}}
        \end{subfigure}
        \begin{subfigure}[t]{0.23\linewidth}
        \includegraphics[width=1.0\linewidth]{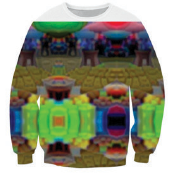}
        \caption{\textsc{Tps}}
        \end{subfigure}
        \begin{subfigure}[t]{0.23\linewidth}
            \includegraphics[width=1.0\linewidth]{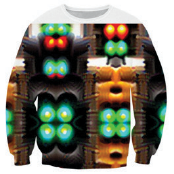}
            \caption{\textsc{E2}-r}
        \end{subfigure}
        \begin{subfigure}[t]{0.23\linewidth}
            \includegraphics[width=1.0\linewidth]{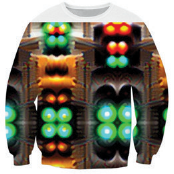}
            \caption{\textsc{E3}-r}
        \end{subfigure}
        \caption{\textbf{Adversarial shirts tested in Section \ref{wearable}}.  Y2 denotes \yt.}
    \label{fig:clothes}
\end{minipage}
\end{figure}

We compute success rates of different patches when tested with YOLOv2 and present the results in Fig.~\ref{fig:pie_dolls}. Comparing across Fig.~\ref{fig:pie_dolls} and the left side of Fig.~\ref{fig:physical_attacks}, we see that paper dolls perform only slightly better than large-format posters.  The performance drop of paper dolls compared to digital simulations, combined with the high fidelity of the paper printouts, leads us to believe that the dominant factor in the performance difference between digital and physical attacks can be attributed to the imaging process, like camera post-processing, pixelation, and compression.

\section{Wearable adversarial examples}  \label{wearable}
Printed posters provide a controlled setting under which to test the real-world transferability of adversarial attacks.  However the success of printed posters does not inform us about whether attacks can survive the complex deformations and textures of real objects.

To experiment with complex real-world transfer, we printed adversarial patterns on shirts using various strategies.  We consider four versions of the \yt patch representing two different scalings of the patch, both with and without boundary reflections to cover the entire shirt (see Figure~\ref{fig:clothes}).
We also consider the \textsc{TPS} patch to see if complex data augmentation can help the attack survive fabric deformations.  Finally, we include the \textsc{Fted}, \enstwo, \ensthree patches to see if these more complex crafting methods facilitate transfer. We collected photos of a person wearing these shirts at ten different locations. For each location and shirt, we took 4 photos with two orientations (front and back) and two distances from the camera. We also took control photos where the person was not wearing an attack. We collected 360 photos in total (see Supple. for a gallery of samples).

We tested the collected images under the same settings as the poster study, and measure the performance of the patches using both AP and success rates. The results are shown in Fig.~\ref{fig:ap_clothes} and Fig.~\ref{fig:success_clothes}.  
We can see that these wearable attacks significantly degrade the performance of detectors.  This effect is most pronounced when measured in AP because, when persons are detected, they tend to generate multiple fragmented boxes. It is also interesting to see that FCOS, which is vulnerable to printed posters, is quite robust with wearable attacks, possibly because shirts more closely resemble the clothing that appears in the training set. When measured in success rates, sweatshirts with \yt patterns achieve $\sim$50\% success rates, yet they do not transfer well to other detectors. Among all \yt shirts, smaller patterns (\ie, \yt-2) perform worse as compared to larger patterns. We also found that tiling/reflecting a patch to cover the whole shirt did not negatively impact performance, even though the patch was not designed for this use.   Finally, we found that augmenting attacks with non-rigid TPS transforms did not improve transferability, and in fact was detrimental.  This seems to be a result of the difficulty of training a patch with such transformations, as the patch also under-performs other patches digitally.

\section{Conclusion}
It is widely believed that fooling detectors is a much harder task than fooling classifiers;  the ensembling effect of thousands of distinct priors, combined with complex texture, lighting, and measurement distortions in the real world, makes detectors naturally robust.
Despite these complexities, the experiments conducted here show that digital attacks can indeed transfer between models, classes, datasets, and also into the real world, although with less reliability than attacks on simple classifiers.

{\scriptsize \noindent \textbf{Acknowledgements}
Thanks to Ross Girshick for his valuable insights into object detectors, and for helping us improve our experiments. This work is partially supported by Facebook AI.}

{\small
\bibliographystyle{splncs04}
\bibliography{reference}
}

\end{document}